\documentclass{article}
\usepackage{arxiv}
\usepackage{amsthm,amsmath,amssymb,amsfonts}
\usepackage[utf8]{inputenc} 
\usepackage{float}
\usepackage[T1]{fontenc}    
\usepackage{hyperref}       
\usepackage{url}            
\usepackage{booktabs}       
\DeclareUnicodeCharacter{2203}{}
\DeclareUnicodeCharacter{211D}{}
\usepackage{amsfonts}     
\usepackage{epsfig,graphicx,subfig}
\usepackage{zref-perpage}
\usepackage{nicefrac}       
\usepackage{microtype}
\usepackage{lipsum}

\usepackage{graphicx}
\usepackage{algorithm, algpseudocode}
\usepackage[table,xcdraw]{xcolor}

\graphicspath{ {./images/} }

\title{Fingerprint Matching using the Onion Peeling Approach and Turning Function}

\author{Nazanin Padkan,$^{\rm a}$\footnote{ nazaninpadkan.ac.ir}
	\ \ B. Sadeghi Bigham,$^{\rm a}$\footnote{ b\_sadeghi\_b@iasbs.ac.ir}
	Mohammad Reza Faraji,$^{\rm a}$\footnote{m.faraji@iasbs.ac.ir} 
	\\$^a$Department of Computer Science and Information Technology, \\Institute for Advanced Studies in  Basic Sciences (IASBS), Zanjan, Iran.
}

\begin{document}
\maketitle
\begin{abstract}
Fingerprint, as one of the most popular and robust biometric traits, can be used in automatic identification and verification systems to identify individuals.
Fingerprint matching is a vital and challenging issue in fingerprint recognition systems. Most fingerprint matching algorithms are minutiae-based. The minutiae in fingerprints can be determined by their discontinuity. Ridge ending and ridge bifurcation are two frequently used minutiae in most fingerprint-based matching algorithms.

This paper presents a new minutiae-based fingerprint matching using the onion peeling approach. In the proposed method, fingerprints are aligned to find the matched minutiae points. Then, the nested convex polygons of matched minutiae points are constructed and the comparison between peer-to-peer polygons is performed by the turning function distance.
Simplicity, accuracy, and low time complexity of the Onion peeling approach are three important factors that make it a standard method for fingerprint matching purposes. The performance of the proposed algorithm is evaluated on the database $FVC2002$. The results show that fingerprints of the same fingers have higher scores than different fingers. Since the fingerprints that the difference between the number of their layers is more than $2$ and the minutiae matching score lower than 0.15 are ignored, the better results are obtained.
\\
\keywords{ Fingerprint Matching \and  Minutiae \and  Convex Layers \and  Turning Function \and Computational Geometry}
\end{abstract}
\newpage
\section{Introduction}

\label{intro}
Due to development in data security systems over the last few decades, the fingerprint is one of the most popular biometric traits.
An automated fingerprint biometric system, depending on the application context, can be a verification or an identification system. While fingerprint verification systems are designed to confirm the identity of individuals, fingerprint identification systems are designed to establish the identity of individuals \cite{wikeclaw2009minutiae}.


A fingerprint matching system takes a fingerprint as an input and compares it with another fingerprint and returns an amount which shows the similarity score (a number between $0$ and $1$) or a binary decision which determines they belong to same finger or not. 

Fingerprint matching approaches are classified into three main groups \cite{maltoni2009handbook}:
\begin{itemize}
\setlength{\itemsep}{1pt}
    \item Minutiae-based matching: Minutiae-based matching is the most frequently used method in fingerprint recognition systems that is based on comparing extracted minutiae points in fingerprint images. The goal is to find the best alignment between two fingerprint that causes the maximum number of matched minutiae points \cite{maltoni2009handbook}. For example, Jain et al. \cite{jain2001fingerprint} proposed a hybrid matching algorithm that uses both minutiae and texture information for matching the fingerprints.

	\item Non-Minutiae feature-based matching:
	When fingerprint images are extremely low-quality, it is difficult to extract minutiae. losing some minutiae or creating some spurious minutiae affect the accuracy of matching system. Extracting ridge pattern (e.g., local orientation and, texture information) in low-quality fingerprint may be more reliably than minutiae, even though their distinctiveness is generally lower \cite{maltoni2009handbook}.  
	
	 Choi et al. \cite{choi2011fingerprint} proposed a new matching algorithm that uses ridge features and minutiae points. Ridge count, ridge length, ridge curvature direction, and ridge type are four ridge features that are considered in their
    studies. Using both minutiae and ridge features increases the performance of recognition against non-linear deformation in fingerprints. Their proposed method is based on a graph matching algorithm. In this method, minutiae points are considered as nodes and ridge feature vectors are the
    edges of the graph.
    
    Correlation-based matching, as a subclass of non-minutiae based matching methods, directly uses the richer gray-scale information of the fingerprint instead of minutiae points \cite{bazen2000correlation}.
    In \cite{nandakumar2004local}, a fingerprint matching algorithm based on local correlation has been proposed. In this method, spatial correlation of regions around the minutiae is used to ascertain the quality of each minutia match.

\end{itemize}

The onion peeling as a computational geometry algorithm is not widely used in computer vision issues such as image retrieval and image matching.
Khazaee and Mohades \cite{khazaei2007fingerprint} established a minutiae-based matching approach which was invariant from translation and rotation. They constructed nested convex polygons of minutiae points. They used the most interior polygon for local matching and proposed another global matching based on convex polygons.
Mazaheri et al. \cite{mazaheri2011fingerprint} proposed a method based on constructed nested convex layers of minutiae and pore points in fingerprint images.

In this paper, a new fingerprint minutiae-based matching approach is proposed which is based on constructed nested convex polygons of matched minutiae. After extracting minutiae points, two fingerprints are aligned to find the matched minutiae points. Then, the nested convex polygons of matched minutiae points are constructed and the comparison
between corresponding polygons is performed by the turning function distance. 
The remaining sections of this paper are organized as follows:
Similarity and distance metric properties are introduced in the Section $2$. Nested Convex Polygons is presented in the Section $3$. a new fingerprint matching algorithm is introduced in the Section $4$. Then, the evaluation results are studied in the Section $5$. Finally, the Conclusion and future work are discussed in the Section $6$.

\section{Similarity and Distance Metric}
One of the most crucial steps of a minutiae-based methods is finding the similarity degree between two minutiae sets. Therefore, similarity and distance metrics are two concepts that need to be defined.
There are many distance metrics that can be used in fingerprint recognition systems. Choosing an appropriate distance metric depends on the application. Shape is an important low-level image feature that can be used in matching purposes such as fingerprint matching. The main challenging issue in shape matching methods is defining a shape representation that is invariant to rotation, translation, scaling, and the curve starting point shift \cite{lee2003similarity}.
In the following, we first define distance metric and similarity metric and then we introduce turning function as an efficient shape matching metric.

\textbf{Distance Metric} 
Given a set $X$, a distance metric is a real-valued function $dist(a,b)$ on the Cartesian product $X*X$ if, for any $a,b,c\in X$, it satisfies the following conditions \cite{chen2009similarity}.

\begin{enumerate}
\item $dist(a,b)\geq 0$,
\item $dist(a,b)= dist(b,a)$,
\item $dist(a,c)\le\ dist(a,b) + dist(b,c)$,
\item $dist(a,b)= 0$ if and only if $a=b$.
\end{enumerate}

\textbf{Similarity Metric}
 Given a set $X$, a similarity metric is a real-valued function $sim(a,b)$ on the Cartesian product $X*X$ if, for any $a,b,c\in X$, it satisfies the following conditions \cite{chen2009similarity}:

\begin{enumerate}
\item $sim(a,b)= sim(y,x)$,
\item $sim(a,a)\geq 0$,
\item $sim(a,a)\geq\ sim(a,b)$,
\item $sim(a,b)+ sim(b,c)\le sim(,z) + sim(y,y)$,
\item $sim(a,a)= sim(b,b)= sim(a,b)$ if and only if $a=b$.
\end{enumerate}

\subsection{Turning Function}
The turning function is an easy and standard tool for representing and describing shapes and in particular polygons. Arkin et al. \cite{arkin1991efficiently} have proposed a method for comparing polygons based on the $L_p$ distance between their turning functions that works for both convex and non-convex polygons. 
The turning function starts from a certain point, called reference, that is arbitrary selected on the shape's boundary \cite{bigham2019survey}. The function measures the counterclockwise tangent as a function of arc s \cite{arkin1991efficiently}. 
$\Theta (0)$ is the angle that reference point makes with x-axis. $\Theta_A(s)$ keeps track of the turning that takes place along the shape, increasing with counterclockwise turns and decreasing with clockwise turns. For a convex polygon, the turning function is monotone and changes in vertices up to $\Theta (0)+2\pi$ and is constant between two consecutive vertices \cite{arkin1991efficiently}.

\begin{figure}
\centering
  \includegraphics[width=0.6\linewidth]{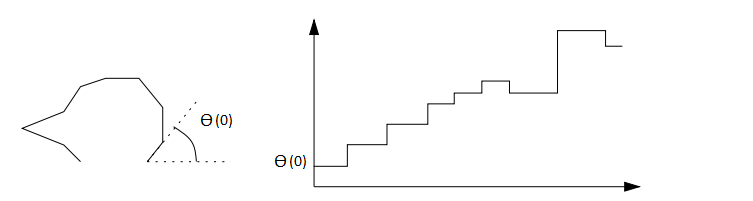}
  \caption{Polygonal curve and turning function \cite{veltkamp2000shape}}
  \label{fig:turn}
\end{figure}

\begin{figure}
\centering
  \includegraphics[width=0.6\linewidth]{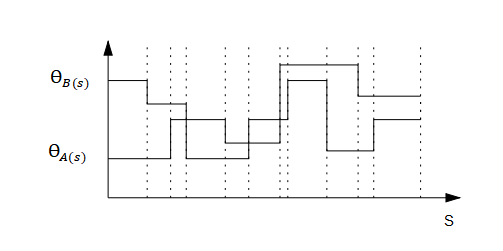}
  \caption{The distance between two polygons is calculated using rectangles enclosed by $\theta_A(s)$, $\theta_B(s)$, and dotted lines \cite{veltkamp2001shape}.}
  \label{fig:dis}
\end{figure}

The distance between polygons $A1$ and $A2$ is defined as $L_p$ distance between their turning functions, see Fig. \ref{fig:dis}.
Consider $\Theta_A1(s)$ and $\Theta_A2(s)$ as two turning functions. The distance between their turning functions is defined as:

\begin{equation}
d(P1,P2) = {\left\| \Theta_P1 - \Theta_P2 \right\|} = ( \int_{0}^1 |\Theta_P1(s) - \Theta_P2(s)|^p ds)^{1/p} 
\end{equation}
Rotating the polygons and choosing different points as the reference point affects the distance. Therefore, the minimum distance over all different reference points and rotation is computed as follows:
\begin{equation}
d(A1,A2) = (\min_{\Theta \in \mathbf{R} , t\in [0,1]}\int_{0}^1 |\Theta_A1(s+v)-\Theta_A2(s)+ \theta |^p ds)^{1/p}
\end{equation}
In the above formula, polygon $A$ is rotated by angle $\theta$ and the reference point is shifted by an amount $v$.
The turning function is invariant to scaling, since the polygons is re-scaled and also invariant under translation and rotation because of having no orientation information \cite{cakmakov2004estimation}. These properties makes it ideal for measuring the similarity between shapes and it can be used in computer vision and specially image retrieval.
According to \cite{arkin1991efficiently}, comparing shapes based on the turning function runs in $O(mn\log{}(mn))$ time where $m$ and $n$ are the number of vertices in two polygons.

\section{Nested Convex Polygons}
\begin{figure}[htb]
\centering
\includegraphics[width=.2\columnwidth]{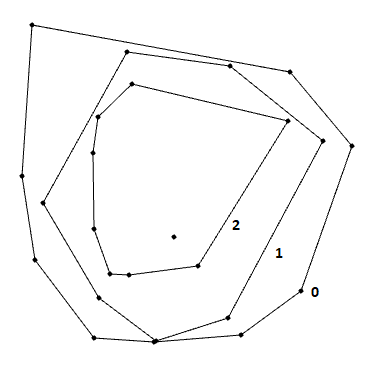} 
\caption{The constructed NCPs of a set point in two-dimensional pace \cite{poulos2004fingerprint}} \label{fig:onion}
\end{figure}

Consider $S = {\{P_1,P_2,...,P_n\}}$ as $n$ points in two dimensional space. The convex layers of $S$ are a sequence of nested convex polygons (NCPs) having the points as their vertices which the outer layer is the convex hull of the points and the rest is formed in a recursive manner (Fig. \ref{fig:onion}). The inner layer may consist of one or two points. This algorithm has also been called "Onion peeling" or "Onion decomposition". There are several algorithm for construction convex hull. Here, Monotone chain \cite{andrew1979another} is iteratively used for constructing nested polygons.

\begin{algorithm}

\caption{Generating the Nested Convex Polygons (NCP) of $n$ points in two-dimensional space \cite{khazaei2007fingerprint}}
\begin{algorithmic}[1]
\label{alg:NCP}
\Require $Points=\{P_1,P_2,...,P_n \}$,Layer={\{\}}, depth=0
\Ensure Nested Convex Polygons of the set $Points$ and depth of each polygon
\While{$N(Points) > 0$}
\State $Layer = \{ \};$
\State $MC = Monotone Chain(S);$
\State $Points=Points-Layer;$
\State $StorePolygonProperties(Layer,depth);$
\State $depth++;$
\EndWhile

\end{algorithmic}
\end{algorithm}

\section{Fingerprint Matching using Turning Function}
\begin{figure}
\centering
  \includegraphics[width=0.6\linewidth]{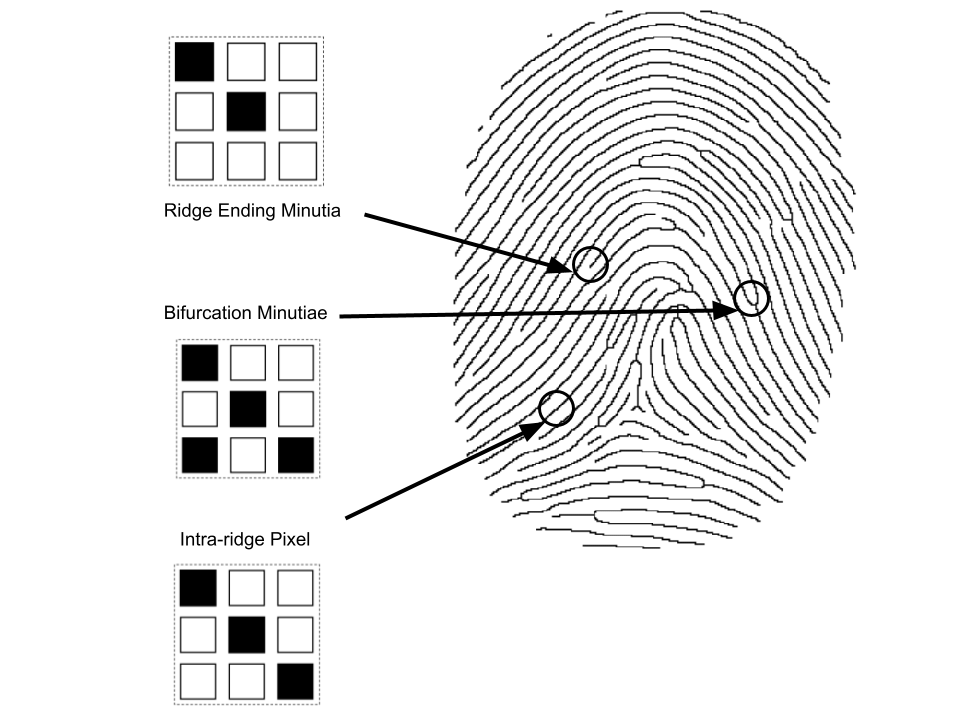}
  \caption[]{The difference between ridge ending, bifurcation minutia, and intra-ridge pixel}
  \label{fig:endbif}
\end{figure}

Since minutia is the most widely used finger features among others, a
strong minutiae extraction algorithm is a vital step in fingerprint identification and verification systems.
Binarization, thinning, and enhancements are three steps are performed on fingerprints in the pre-processing process.
The number of neighboring pixels around a pixel on a thinned fingerprint is used to detecting bifurcation and ending minutiae points . Pixels which have a single neighbor is selected as termination, and if they have more than two neighbors, they are considered as bifurcation points \cite{bansal2011minutiae}. A pixel that has two neighbors is an intra-ridge pixel and not regarded as minutiae (Figure \ref{fig:endbif}) .

The proposed algorithm is divided into two processes. In the first process, minutiae points are extracted using MinuE(). In this step, fake minutiae are also ignored. Minutiae are considered as triplets $m=\{(x,y),\theta, type\}$ with three elements which are the minutiae location, orientation, and type. The type can be ridge ending or ridge bifurcation. Then, minutiae pairs are found in the input and template fingerprints using MM(). In the next step, the nested convex polygons (NCP) of matched minutiae points are constructed and the distance between corresponded turning functions is computed. \newline

\begin{algorithm}
\caption{The Proposed Fingerprint Matching Algorithm}
\begin{algorithmic}[1]
\Require Input and template fingerprint images
\Ensure Matching score over the average of turning distances and minutiae matching
\State Extract minutiae points of the input and template fingerprints:
   \State $I$ =     \textbf{MinuE}(Input);
   \State $T$ =                 \textbf{MinuE}(Template);
\State Find matched minutiae points of the input and template fingerprints:
\For{i\; in I}
    \For{j\; in T}
        \State $I_k$ , $T_k$ = \textbf{MM($I$,$T$)};
    \EndFor
\EndFor
\State Construct the NCPs of matched minutiae points:
\For{i\; in $I_k$}
    \State $L_i$ = \textbf{NCP($I_k$)};
\EndFor
\For{j\; in $T_k$}
    \State $L_j$ = \textbf{NCP($T_k$)};
\EndFor
\State Compute the turning functions of the NCPs in the input and template fingerprints:
\State $TF1$ = \textbf{TF($L_i$)};
\State $TF2$ = \textbf{TF($L_j$)};
\State Compute the average of distances between \textbf{TD($TF1$,$TF2$)};
\State Compute the similarity score.

\end{algorithmic}
\end{algorithm}

\subsection{Minutiae Matching}
Consider ${I=\{m_1,m_2,...,m_{\rm{\emph{m}}}\}}$ and ${T=\{m'_1,m'_2,...,m'_{\rm{\emph{n}}}\}}$ as the representations of the input and template fingerprint images. These representations are feature vectors of fingerprint minutiae points.
Minutiae $m_i$ in the input fingerprint and minutiae $m'_j$ in the template fingerprint will be matched, if their  orientation difference $D$ and spatial distance $S$ are smaller than specific tolerances $r_0$ and $\theta_0$, respectively: 

\begin{equation}
sd(m'_j,m_i) =  \sqrt{(x'_j - x_i)^2 + (y'_j - y_i)^2} \leq r_0
\end{equation}

\begin{equation}
dd(m'_j,m_i) =  \min(\left| \theta'_j - \theta_i \right|, 360^{\circ}- \left| \theta'_j - \theta_i \right|) \leq \theta_0
\end{equation}

The aim is to maximize the number of correct matched minutiae. So, aligning the input and template fingerprints is an obligatory step. Consider (s,$\Delta\theta$,$\Delta{x}$,$\Delta{y}$) is the set of transformations that is needed to be estimated to find the best matching alignment:\newline

\begin{equation}
F_{s,\Delta\theta,\Delta{x},\Delta{y}}
   \begin{bmatrix} 
   x_{temp} \\
   y_{temp} \\
   \end{bmatrix} =
   s\begin{bmatrix} 
   \cos\Delta\theta & -\sin\Delta\theta  \\
   \sin\Delta\theta & \cos\Delta\theta  \\
   \end{bmatrix} 
   \begin{bmatrix} 
   x_{input} \\
   y_{input} \\ 
   \end{bmatrix} +
    \begin{bmatrix} 
   \Delta{x} \\
   \Delta{y} \\ 
   \end{bmatrix}.
\end{equation}
\begin{figure}[htb]
\centering
\includegraphics[width=.5\columnwidth]{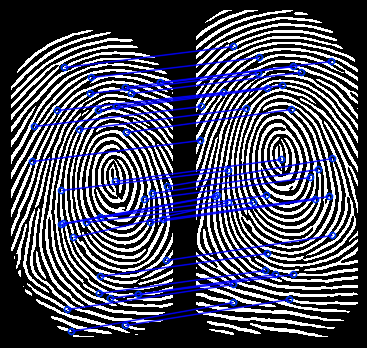} 
\caption{Implementation of minutiae matching on two fingerprints of the same finger from the $FVC2002$ db2. } \label{fig:fingermatching}
\end{figure}

There is a traditional way to calculate the efficiency of a minutiae-based matching system which is used for the first step of matching algorithm:
\begin{equation}
Minutiae \; \;score = \frac{k}{(m+n)/2}
\end{equation}
where $k$ is the number of matched minutiae on two fingerprint images and $m$ and $n$ represent the number of minutiae on the input and template fingerprint images.

\subsection{Nested Convex Polygons of Matched Minutiae Points}

\begin{figure}
    \centering
    \subfloat[]{{\includegraphics[width=.2\columnwidth]{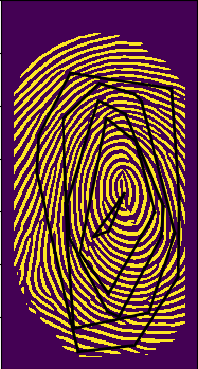}}}%
    \qquad
    \subfloat{{\includegraphics[width=.2\columnwidth]{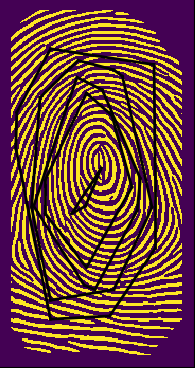}}}%
    \caption{The nested convex polygons of minutiae points in two fingerprints of same finger.}%
    \label{fig:onionpolygon}%
\end{figure}

After detecting the matched minutiae points, their NCPs are constructed in the fingerprints.
Let ${I_k=\{m_1,m_2,...,m_{\rm{\emph{k}}} \}}$ and ${T_k=\{m'_1,m'_2,...,m'_{\rm{\emph{k}}} \}}$ be the sets of matched minutiae points on the input and template fingerprint images.
Using the onion peeling approach, the nested convex polygons are constructed for the sets $I_k$ and $T_k$ (Figure \ref{fig:onionpolygon}). The depth and vertices of each polygon are stored.

when the nested polygons are constructed, it is time to compare the peer-to-peer polygons. Turning function TF() is a standard tool that is used here for representing and describing nested convex polygons of minutiae points.
After creating all the turning functions of the nested convex polygons, it needed to check the distance in the optimal orientation between the two peer-to-peer convex polygons using TD(). 

let $S={\{A_1,A_2,...,A_l\}}$ be the set of optimal turning distances of layers and $l$ be the number of matched layers of the fingerprint images. In the case of varying the number of layers in the input and template fingerprint images, the minimum number of them is selected as $l$ and comparison is done from the most inner layer. The total average of turning distance of the input and template fingerprint images is computed as:

 \begin{equation}
     Average = \frac{ \sum_{i=1}^{l}A}{l}    
 \end{equation}

The average of the turning function distances of the NCPs is not adequate for comparing fingerprints.
Since the number of matched minutiae in fingerprints of the same finger is more than those in different fingers, the number of nested convex polygons are also higher. 

Since the algorithm tries to align fingers and find the optimal orientation between peer-to-peer layers, it is expected to have more scores for different fingers with a few uncomplicated convex nested polygons (usually two or one layers). On the other hand, high resolutions fingerprints of the same finger may be considered as non-match pairs because of the higher number of complicated onion layers of matched minutiae and the score will increase accumulatively.  
So, it is obligatory to relate the number of matched minutiae to the average of turning function to have a better score for comparing fingerprints. Minutiae score shows the similarity whereas Average shows the distance. Therefore, division is a good way to relate them.

 \begin{equation}
    \alpha = \frac{Minutiae \; \;score }{Average}
 \end{equation}

The amount $\alpha$ is $0$ and $\infty$ and it is scaled to an amount between $0$ and $1$:

\begin{equation}
    Final\; Score = 1 - 2^{-\alpha}
\end{equation}

\subsection{System Evaluation}

The result of comparing the input and template fingerprint images is an amount between $0$ and $1$ called "matching score" that shows the similarity between those images \cite{maltoni2009handbook}. Closer the score to $1$ shows more similarity between two fingerprint images. The decision of considering two fingerprints as matching or non-matching pairs is defined by regulating a threshold. 
Two fingerprints are considered as matching pairs if the matching score is higher than or equal to the threshold $t$ and fingerprints which their matching score is lower than $t$ are regarded as non-matching pairs. 

If the input fingerprint is matched to an enrolled fingerprint, this pair is called a genuine match. When the comparison is between samples from different users, this is called an impostor match \cite{dunstone2008biometric}.

A varying score thresholds are applied to the similarity scores, so $FNMR$ and $FMR$ curves can be calculated. To determine $FMR$, all possible comparisons of impostor fingerprint pairs are performed and the number of all impostor pairs incorrectly classified as a genuine pair are calculated \cite{dunstone2008biometric}. To calculate $FNMR$, all possible comparisons of genuine fingerprint pairs are performed and only those which were incorrectly classified as an impostor pair are detected.

If the threshold is increased, $FNMR$ is increasing and as a result, $FMR$ is decreased. Increasing the threshold makes the system more secure while the system is more tolerant by decreasing the threshold. The cross point of $FMR$ and $FNMR$ curves is called Equal Error Rate ($EER$) that is the best value for different thresholds \cite{maltoni2009handbook}.


The performance of the proposed algorithm is also evaluated using Accuracy (ACC), Precision (PR), Recall (RC), and F-measure (F) that are going to be introduced. 
The pairs that are correctly and incorrectly recognized by the algorithm are termed as True Positive (TP) and True Negative (TN), respectively. On the other hand, false positive (FP) is genuine pair that is wrongly declared fake and false negative (FN) is the pair that fake wrongly assigned to genuine. \\

\begin{center}
    PR = TP / (TP + FP) \\
    RC = TP / (TP + FN) \\
    ACC = (TP + TN) / (TP + TN + FP + FN) \\
    F = (2 $\times$ PR $\times$ RC)/(PR + RC))
    
\end{center}

\begin{align*}
 Precision &= \frac{\text{Number of relevant images retrieved}}{\text{Total number of images retrieved}},
\end{align*}

\subsection{Experimental Results}
\begin{table}[]
\centering
\caption{Some results of implementing the proposed algorithm to different fingers of the $FVC2002$ $db2_B$.}
\begin{tabular}{|c|c|c|c|c|}
\hline
\textbf{\begin{tabular}[c]{@{}c@{}}Fingerprint\\ Names\end{tabular}} & \textbf{\begin{tabular}[c]{@{}c@{}}Minutiae\\ Matching \\ Score\end{tabular}} & \textbf{\begin{tabular}[c]{@{}c@{}}the number\\  of \\ convex polygons\end{tabular}} & \textbf{average} & \multicolumn{1}{l|}{\textbf{\begin{tabular}[c]{@{}l@{}}Final\\ Score\end{tabular}}} \\ \hline
\begin{tabular}[c]{@{}c@{}}101\_1\\ 101\_2\end{tabular} & 0.44 & \begin{tabular}[c]{@{}c@{}}5\\ 5\end{tabular} & 26.78 & 0.98 \\ \hline
\begin{tabular}[c]{@{}c@{}}101\_1\\ 101\_3\end{tabular} & 0.33 & \begin{tabular}[c]{@{}c@{}}3\\ 4\end{tabular} & 52.97 & 0.75 \\ \hline
\begin{tabular}[c]{@{}c@{}}101\_5\\ 101\_6\end{tabular} & 43.54 & \begin{tabular}[c]{@{}c@{}}3\\ 3\end{tabular} & 43.54 & 0.77 \\ \hline
\begin{tabular}[c]{@{}c@{}}103\_2\\ 103\_1\end{tabular} & 0.38 & \begin{tabular}[c]{@{}c@{}}4\\ 4\end{tabular} & 45.68 & 0.88 \\ \hline
\begin{tabular}[c]{@{}c@{}}103\_6\\ 103\_7\end{tabular} & 0.17 & \begin{tabular}[c]{@{}c@{}}3\\ 2\end{tabular} & 33.3 & 0.73 \\ \hline
\begin{tabular}[c]{@{}c@{}}105\_1\\ 105\_4\end{tabular} & 0.25 & \begin{tabular}[c]{@{}c@{}}2\\ 2\end{tabular} & 73.2 & 0.57 \\ \hline
\begin{tabular}[c]{@{}c@{}}107\_3\\ 107\_6\end{tabular} & 0.26 & \begin{tabular}[c]{@{}c@{}}3\\ 3\end{tabular} & 63.53 & 0.65 \\ \hline
\end{tabular}
\end{table}

This section describes the experiments for evaluating the proposed algorithm.
$FVC2002$ database is used here to carry out the experiments of the method. $FVC2002$ has four different databases (DB1, DB2, DB3, and DB4). These databases are collected using different sensors and
technologies. The quality of DB1 images is higher than the others. DB3 images are of low quality and DB2 and DB4 images are of moderate quality. Each minor database has 80 fingerprints of 10 individuals (8 images per individual). 
The proposed method is evaluated using Accuracy, Precision, Recall, and F-measure on $FVC2002$ $DB2\_b$. Biometric error rates such as $FMR$ and $FNMR$ are also needed to evaluate how accurate the proposed method is. Some of the comparisons are shown in Table $1$ and Table $2$. The compared fingerprints have different qualities (low, medium, and good) to get more accurate results. 

As can be seen, fingerprints of the same finger (Table $1$) have higher minutiae matching score in comparison with those from different fingers (Table $2$) and as a result, the number of nested convex layers is also higher. The comparison of the last two columns of tables shows a higher score in Table $1$. Regulating an appropriate threshold $t$ is a vital step that should be regarded. Fingerprints which their score is higher than $t$ are regarded as matched and those with lower score considered as non-match pairs.

There are some parameters and thresholds that can be regulated and affect the performance of the method:

\begin{enumerate}
    \item In the post-processing stage of extracting minutiae, minutiae that are within a certain distance $rm$ of each other are merged. Changing $rm$ affects the number of final minutiae points and as a result the number of nested convex polygons are affected.
    \item Fingerprints which have minutiae matching score lower than a specific amount $sim$ are rejected.
    \item The results of the proposed method show that the number of NCPs in the same fingerprints are equal and in many cases, one of them has one more layer. The difference between the number of nested convex polygon NCPs $diff$ can be regarded as a matching factor.
    \item In the process of matching minutiae, distance $d$ and orientation $th$ thresholds are two important factors that can be regulated.  
\end{enumerate}

According to the Plots in Figures, $7$ and $8$, applying differences between the layers has a positive effect on the results. Since the difference in the number of layers of the same fingerprint is usually not large, it is possible to avoid comparing fingerprints that have a difference of more than $2$ layers. It is also possible to avoid comparing fingerprints that have a very small miniature matching rate (for example, under $15$ percent).

Comparison between four plots in Figures $7$ and $8$ shows that fewer error rates are obtained when $sim=15$, $diff=2$, $rm=5$, $th=10$, and $d=15$. It means that fingerprints with minutiae matching score lower than $0.15$ and differences of more than two layers are ignored. The other three parameters do not have a significant effect. In this case, EER is also lower than the other three plots (EER is about $0.23$ and the threshold is 0.41). Besides, It has higher Accuracy (0.85) and F-measure (0.3) in comparison with others (Figure 8). The precision and recall are 0.31 and the best threshold for comparing fingerprints is about $0.72$.

\begin{table}[]
\centering
\caption{Some results of implementing the proposed algorithm to different fingers of $FVC2002$ $DB2\_b$.}
\begin{tabular}{|c|c|c|c|c|}
\hline
\textbf{\begin{tabular}[c]{@{}c@{}}Fingerprint\\ Names\end{tabular}} & \textbf{\begin{tabular}[c]{@{}c@{}}Minutiae\\ Matching\\ Score\end{tabular}} & \textbf{\begin{tabular}[c]{@{}c@{}}The Number \\ of\\ Convex polygons\end{tabular}} & \textbf{Average} & \textbf{\begin{tabular}[c]{@{}c@{}}Final\\ Score\end{tabular}} \\ \hline
\begin{tabular}[c]{@{}c@{}}104\_2\\ 110\_1\end{tabular} & 0.16 & \begin{tabular}[c]{@{}c@{}}2\\ 2\end{tabular} & 0.146 & 0.54 \\ \hline
\begin{tabular}[c]{@{}c@{}}109\_8\\ 102\_7\end{tabular} & 0.10 & \begin{tabular}[c]{@{}c@{}}2\\ 3\end{tabular} & 0.20 & 0.31 \\ \hline
\begin{tabular}[c]{@{}c@{}}109\_5\\ 107\_7\end{tabular} & 0.12 & \begin{tabular}[c]{@{}c@{}}1\\ 1\end{tabular} & 0.11 & 0.51 \\ \hline
\begin{tabular}[c]{@{}c@{}}108\_6\\ 110\_4\end{tabular} & 0.13 & \begin{tabular}[c]{@{}c@{}}1\\ 1\end{tabular} & 0.13 & 0.49 \\ \hline
\begin{tabular}[c]{@{}c@{}}102\_1\\ 101\_1\end{tabular} & 0.15 & \begin{tabular}[c]{@{}c@{}}2\\ 2\end{tabular} & 0.24 & 0.36 \\ \hline
\begin{tabular}[c]{@{}c@{}}103\_6\\ 102\_7\end{tabular} & 0.16 & \begin{tabular}[c]{@{}c@{}}2\\ 2\end{tabular} & 0.23 & 0.39 \\ \hline
\begin{tabular}[c]{@{}c@{}}104\_5\\ 103\_3\end{tabular} & 0.13 & \begin{tabular}[c]{@{}c@{}}2\\ 2\end{tabular} & 0.147 & 0.49 \\ \hline
\begin{tabular}[c]{@{}c@{}}110\_4\\ 101\_1\end{tabular} & 0.21 & \begin{tabular}[c]{@{}c@{}}3\\ 3\end{tabular} & 0.19 & 0.53 \\ \hline
\end{tabular}
\end{table}

\begin{figure}[ht] 
\centering
  \label{fig:fm} 
  \begin{minipage}[b]{0.6\linewidth}
    \includegraphics[width=.6\linewidth]{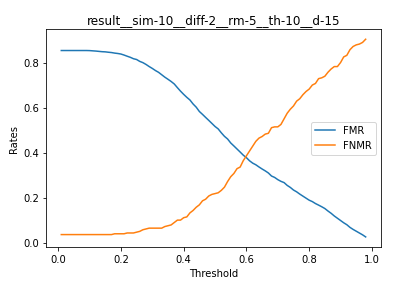} 
    \vspace{4ex}
  \end{minipage}
  \begin{minipage}[b]{0.6\linewidth}
    \includegraphics[width=.6\linewidth]{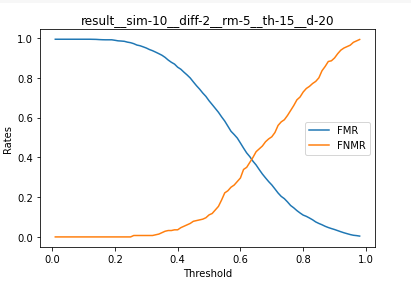} 
    \vspace{4ex}
  \end{minipage} 
  \begin{minipage}[b]{0.6\linewidth}
    \includegraphics[width=.6\linewidth]{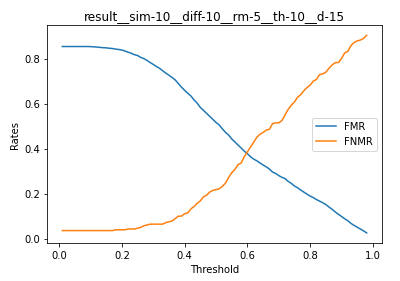} 
    \vspace{4ex}
  \end{minipage}
  \begin{minipage}[b]{0.6\linewidth}
    \includegraphics[width=.6\linewidth]{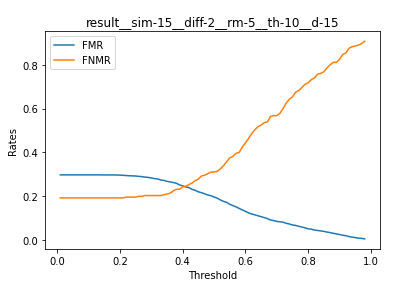} 
    \vspace{4ex}
  \end{minipage} 
  \caption{False Match Rate ($FMR$) and False Non-Match Rate ($FNMR$) on $DB2\_b$, $FVC2002$.} 
\end{figure}


\begin{figure}[ht] 
\centering
  \label{fig:prf} 
  \begin{minipage}[b]{0.6\linewidth}
    \includegraphics[width=.6\linewidth]{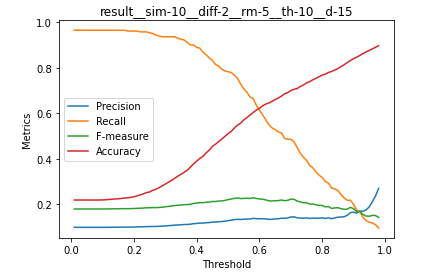} 
    \vspace{4ex}
  \end{minipage}
  \begin{minipage}[b]{0.6\linewidth}
    \includegraphics[width=.6\linewidth]{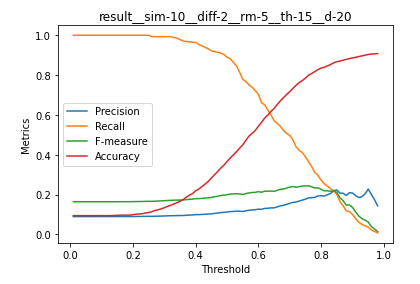} 
    \vspace{4ex}
  \end{minipage} 
  \begin{minipage}[b]{0.6\linewidth}
    \includegraphics[width=.6\linewidth]{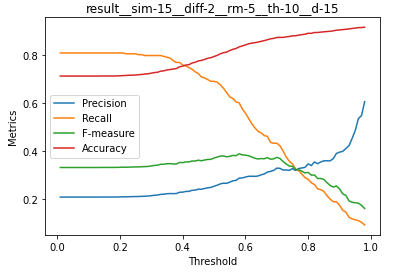} 
    \vspace{4ex}
  \end{minipage}
  \begin{minipage}[b]{0.6\linewidth}
    \includegraphics[width=.6\linewidth]{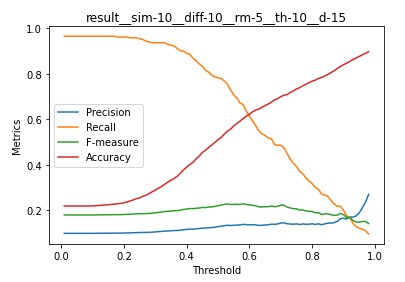} 
    \vspace{4ex}
  \end{minipage} 
  \caption{Precision, Recall, F-Measure, and Accuracy on $DB2\_b$, $FVC2002$.} 
\end{figure}

\clearpage

\section{Conclusion and Future Work}
In this paper, a new minutiae-based fingerprint matching algorithm that utilizes nested convex polygons and the turning function is proposed. 
Construction of the NCPs of the matched minutiae and then, comparing them with the turning function distance have low time complexity. Also, the turning function distance is invariant to rotation, scaling, and translation which is a useful way for matching polygons. 

The results show that the proposed algorithm is efficient for matching fingerprints. The score of matching fingerprints of the same finger is approximately higher than the score of different fingers. The best results are obtained when fingerprints with minutiae matching score lower than $0.15$ and layer differences more than $2$ are ignored. In this case, the best threshold for comparing fingerprints is $0.72$ and accuracy and precision are 0.85 and 0.31, respectively.

In future works, other fingerprint details in addition to minutiae are used to create a more accurate method. Using classification methods like Support Vector Machine (SVM) can be used for classifying fingerprints that reduce the matching time. It is recommended to use a more robust method for enhancing fingerprints and extracting minutiae points such as deep convolutional networks which can significant effects on noisy and low-quality fingerprints. Besides, the nested convex polygons can be constructed for bifurcation and ending minutiae separately.

\bibliography{export}

\end{document}